\documentclass[11pt,a4paper]{article}
\usepackage{authblk}
\usepackage[hyperref]{acl2021}
\usepackage{times}
\usepackage{latexsym}
\usepackage{xspace}
\usepackage[ruled,vlined]{algorithm2e}

\usepackage{microtype}
\usepackage{graphicx}
\usepackage{tabularx}
\usepackage{booktabs}
\usepackage{bm}

\aclfinalcopy 

\title{Go Forth and Prosper: Language Modeling with Ancient Textual History}

\author[1]{Rik Koncel-Kedziorski}
\author[1,2]{Noah A. Smith}
\affil[1]{University of Washington}
\affil[2]{Allen Institute for AI}
\affil[ ]{\texttt{kedzior@uw.edu, nasmith@cs.washington.edu}}

\date{}

\begin{document}
\maketitle

\begin{abstract}
We introduce a technique for improving document-level language models (LM) by leveraging ``ancient history'':  text that is outside the LM's current context window. 
We learn an auxiliary function to select spans from the ancient history which can help the LM to predict future text. 
The selected text spans are then copied directly into the LM's context window, replacing less predictive spans.  
This method can improve perplexity of pretrained LMs with no updates to the LM's own parameters. 
We further observe that an auxiliary function trained in a specific textual domain like Wikipedia will also work in a substantially different domain such as scientific publications. 
With this technique we see a 7\% perplexity reduction on Wikipedia articles, and a 12\% perplexity reduction on scientific texts. 
\end{abstract}

\section{Introduction}
Modern language models (LMs) make use of increasingly long contexts.
Architectures which attend to upwards of 3,000 tokens have been proposed, but many current pretrained models use a context window of 512 or 1,024 tokens. 
Yet a great deal of human communication---and thus of LM applications---consists of long-form text. 
Scientific, technical, medical, and legal knowledge is often conveyed via documents tens of thousands of tokens long. 
To effectively use such material, LMs will need to consume long contexts. 

In this paper we introduce a simple technique for improving language modeling of long documents by effectively extending the LM's accessible history beyond the architecture-specified context window and into the ``ancient history''---text which comes before the beginning of the context window. 
We train an auxiliary function to select the parts of the ancient history that are most predictive of the future text. 
The conditioning context of the LM is then altered to include these predictive parts of the ancient history. 

An important advantage of this technique is that it can be used with off-the-shelf pretrained LMs with no additional tuning of the LM's parameters. 
This quality of our technique is especially relevant given the trend toward keeping LM parameters secret from both researchers and public, as has been done with GPT3 \citep{brown2020language}.
As the computational and social costs of language modeling show no signs of abating, it is reasonable to anticipate that future large LMs will also be similarly inaccessible. 

We apply the proposed technique to language modeling with the popular GPT2 family of models \citep{radford2019language}. 
We observe perplexity reductions on both in-domain and cross-domain language modeling data across model sizes. 
We also find that we can train the ancient history selection function in one domain and apply it successfully in another, evidencing for the transferability of the proposed technique.\footnote{Code available at \url{https://github.com/rikdz/AHLM}}





\section{Method}
A language model defines a probability distribution over sequences of words $\boldsymbol{w} = \langle w_0, \ldots, w_n \rangle$, usually factorized as
\begin{equation}
    p_{\mathit{LM}}(\boldsymbol{w}) = \prod_{t} p_{\mathit{LM}}(w_t \mid \boldsymbol{w}_{0:t-1}).
\end{equation}
Due to computational limitations, the conditioning context is bounded to (at most) $k$ tokens.  Pre-neural language models fixed $k$ at values like 2 or 4; current research often uses values of 512 or 1,024, sometimes as high as 3,000.  The classical assumption is that the most recent words are the most useful, i.e., the model should define $p_{\mathit{LM}}(w_t \mid \boldsymbol{w}_{t-k:t-1})$.  When modeling longer texts, and with longer contexts, it is not clear that this assumption is correct.  Our approach aims to condition on both the most recent history as well as automatically selected content from the arbitrarily distant past.  Further, we do so without modifying the underlying language model's parameters.

\begin{figure*}[ht]
\includegraphics[width=\textwidth]{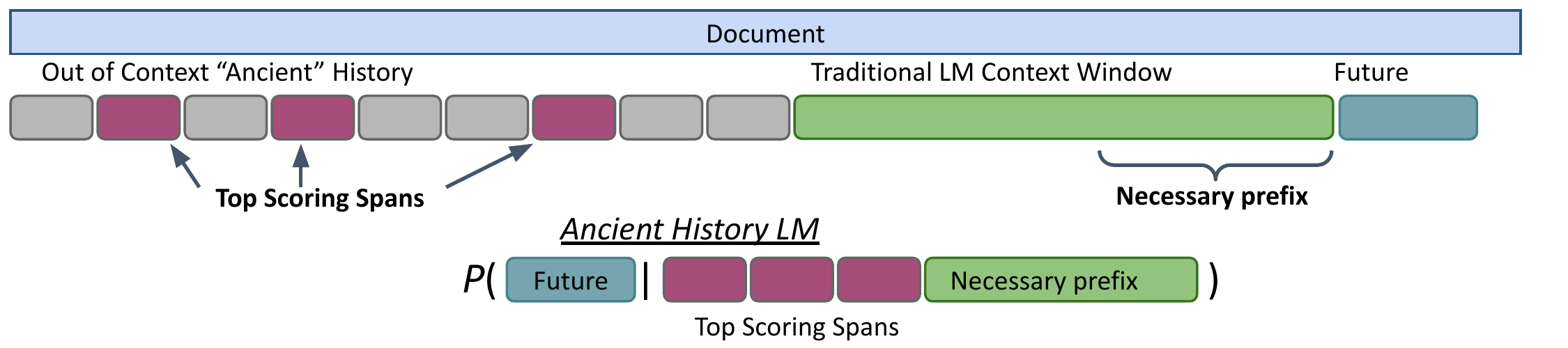}
\caption{Overview of our approach. A traditional LM context window only covers part of the document. Our method brings the best spans from the ancient history into the context window for improved modeling of future text.}
\label{fig-overview}
\end{figure*}

To do this, we divide the $k$-word context into two parts.  The first $j$ positions, conventionally filled by history words at positions $t-k$ through $t-k+j$, will be filled by \textbf{selected} content from anywhere in the full history, which we will denote $\boldsymbol{s}^t_{1:j}$.    The remaining $k-j$ positions are unchanged (i.e., filled by $\boldsymbol{w}_{t-k+j:t-1}$, which we call the \textbf{necessary prefix}).  Figure~\ref{fig-overview} illustrates the high-level idea.  Note that new selected history content is selected separately for each predicted word.\footnote{For sliding window inference, selected history is populated only once per window.}

To apply our approach at inference time, the only modification to querying an existing LM in the conventional way is to replace the first $j$ positions of each word's history with selected tokens.  To accomplish this, we need a selection function.

\paragraph{Selection Function}
\label{sec-selection}
The selection function looks into the out-of-context (``ancient'') history and selects spans which seem important for understanding the document. 
A strong selection function will consider the necessary prefix, $\boldsymbol{w}_{t-k+j:t-1}$, and select content from anywhere in the ancient history $\boldsymbol{w}_{0:t-k+j-1}$ to add to the context in the query to the LM.  
In this work, we identify a span size $\ell$, such that $\ell \leq j$ and $j \equiv 0 \pmod \ell$.   We break the ancient history into overlapping $\ell$-length spans and score them with the selection function; only the $j/\ell$ top-scoring spans will be included in the word history. Future work might consider variable-length spans, or determine the total amount of selected content (i.e., $j$) per document or timestep. 

The scoring function is an auxiliary neural network that is trained on text.  Its training  objective function is designed to predict the log ratio of the likelihood of \emph{future} words given the selected span, versus the likelihood given only the necessary prefix: 
\begin{equation}
\ln\left(
    \frac{p_{\mathit{LM}}(\boldsymbol{w}_{t:t+s} \mid \boldsymbol{s}^t_{1:j}, \boldsymbol{w}_{t-k+j:t-1})}{p_{\mathit{LM}}(\boldsymbol{w}_{t:t+s} \mid \boldsymbol{w}_{t-k+j:t-1})}
\right)
    \label{eq:ratio}
\end{equation}
We therefore train the auxiliary network as a regression model, i.e., minimizing the squared error between the score it assigns to a candidate span and the quantity in Eq.~\ref{eq:ratio}.  Training examples are easily constructed from a text corpus; we sample a collection of span/prefix pairs from the text and approximate the probabilities in Eq.~\ref{eq:ratio} using a pretrained language model (GPT2 large).  

Though there are many choices for the selection function's architecture, we opt to finetune a pretrained language model. 
We encode the span and prefix jointly with GPT2 small. 
We then pass the last layer of the final token through a two-layer feedforward network with ReLU activation that predicts the score.
We note that this design choice is suboptimal for many applications due to its computational cost. 
Future work could reduce the time complexity of our technique by reusing previously computed representations or modeling the selection function with a lighter architecture. 

\paragraph{Oracles}
An upper bound for the proposed method is given by an {\bf oracle} selection function which uses the probabilities computed in Eq.~\ref{eq:ratio}, with the actual future words, to select the optimal history for each $w_{t}$ from among the candidates. 
When run on training data, this oracle can be used to select the two  hyperparameters introduced above: the total length of the selected content $j$ and the span length $\ell$.

\section{Experiments}

We conduct experiments to determine the effect of incorporating ancient history into the context windows of pretrained LMs.
Specifically, we explore the impact of the selector function on in-domain and out-of-domain data. 

\paragraph{Experimental Setup}
Our experiments follow the typical language modeling evaluation setup with one modification: rather than concatenating all documents together, we reset the context window between documents. 
This means that the beginning of documents will have less context than in standard language modeling setups.
For this reason, our baseline scores are slightly different from other reported results. 
We compute average token perplexity across each dataset. 
We use a sliding window of 1024 tokens with a stride of 256 tokens during inference and repopulate the selected history each stride. 

To determine the hyperparameters $j$ and $\ell$, we run a grid search over oracles with parameters $j \in \{256,768,512,128\} \times \ell \in \{8,16,32,64,128\}$.
We find $j=512$, $\ell=64$ works best with oracles on training data and use these in our model.
For training the selector function, we calculate the scores from Equation~\ref{eq:ratio} for 728K span/prefix pairs from the Wikitext-2 training dataset.
The model is optimized with AdamW with a learning rate of $0.00001$.

In our experiments, we compare our ancient history approach to a baseline model that follows the convention of always using the $k$ most recent tokens as the history.  
\subsection{In-Domain Experiment}
We first consider the GPT2 large model on Wikitext-2.
We compare a standard context window ($k$ most recent words) to our ancient history selection method, where 8 spans of $\ell=64$ tokens are drawn from earlier in the article with our trained selection function (\S\ref{sec-selection}). 
This result is shown in the left half of Table~\ref{tab-res}, line 1.
We observe a 0.9 absolute perplexity reduction. 

One question that arises is the extent to which the selection function is essentially learning the language modeling task.  We evaluate an oracle selection function that chooses ancient history spans that have the highest overlap with the future words.  Despite having access to the text to be predicted, this approach only outperforms the baseline by 0.1.  
We take this as evidence that our selection function is learning a notion of ``useful'' ancient history, not merely to predict the words that will appear in the future (the language modeling task).

Line 2 of Table~\ref{tab-res} shows bigger performance gain when using ancient history in the off-the-shelf GPT2 small language model. 
As smaller models are often used where inference latency is important, future work could investigate selection functions with lower latency or compute costs. 

On Line 3 of Table~\ref{tab-res} we see the impact of finetuning. 
Finetuning to the in-domain data strongly improves model perplexity compared to the off-the-shelf GPT2 small model. Yet even the finetuned model benefits from the use of ancient history contexts.\footnote{The GPT models are trained with contiguous text, and so the non-contiguous selected history is slightly ``out of domain.'' 
We also tried finetuning the LM on oracle selected history data, but the results were similar to the traditionally finetuned model.}

Line 4 shows the perplexity obtained by replacing our selection function with an oracle selection function (i.e., the one used to select hyperparameter values). 
We see that our learned function comes within 0.1 perplexity of this oracle.

\subsection{Cross-Domain Experiment}
To understand how the model performs over longer documents and on out-of-domain data, we evaluate on 1,451 scientific papers from the S2ORC corpus \citep{lo-etal-2020-s2orc}.  
Documents in this corpus are on average twice as long as Wikipedia articles (6K vs.~3K words), and cover complex scientific topics in greater detail, a challenge for LMs. 

We see similar improvements on the S2ORC subset, shown on the right side of Table~\ref{tab-res}.
The selection function used here was trained only on the Wikitext2 training data, indicating that a selection function trained once may be applied across domains. 
It is notable that the differences between the baselines and proposed technique are larger in this cross-domain setting. 
This may be due to the fact that documents are longer in the science domain, so that there is ofte more useful information in the out of context document history. 
Additionally, the language used in scientific texts is rather different from the training data of GPT2.
The selection function is less affected than a language model by such vocabulary mismatches, perhaps because selecting spans from a document is easier than choosing from a large, long-tailed vocabulary. 

\paragraph{Example}
To illustrate how ancient history is used, we consider the modeling of the 2016 CVPR paper ``Convolutional Two-Stream Network Fusion for Video Action Recognition'' \citep{Feichtenhofer2016ConvolutionalTN}, which appears in the S2ORC corpus. 
This paper studies fusion strategies in ConvNet architectures for the task of recognizing human actions in video. 
Among the top 20 highest-impact\footnote{Impact is measured by perplexity difference observed when using selected versus original context.} terms included by the selection function and absent from the original context window, we find {\it temporal, -cite-} (a dataset specific replacement for citation markers), {\it spatiotemporal, training, layers,} and {\it ConvNet}.
The most frequently selected span from the ancient history -- appearing in 8 context windows -- is ``{\it \ldots individual frames can be ambiguous, and motion cues are necessary.
Consider, for example, discriminating walking from running, yawning from laughing, or in swimming, crawl from breast-stroke.
The two-stream architecture -cite- incorporates motion information by training separate ConvNets for both appearance \ldots}''.
A single sentence which appears in over half of all selected histories is ``{\it We build upon the the two-stream architecture in -cite-}'', and additional variants of this sentence are found in other frequently selected windows (such as the above).
These spans demonstrate the summary-like quality of selected history, but notably do not contain the most informative single word ({\it temporal}). 
This indicates that the selection function often chooses content from the ancient history that will be useful to include with a specific necessary prefix, rather what might be best for modeling the document as a whole. 


\begin{table*}
\begin{tabularx}{\textwidth}{llXrrr|rr}
    \toprule
    &&& \multicolumn{3}{c}{Wikitext-2 val.~ppl. ($\downarrow$)} & \multicolumn{2}{c}{ S2ORC val.~ppl. ($\downarrow$)} \\
     & model &&  original & overlap oracle & AHLM & original & AHLM \\ \midrule 
     1.&GPT2 Large && 16.56 & \emph{16.43} & \bf 15.64 & 14.12 & \bf 12.62 \\ \midrule  
    2.&GPT2 Small  && 26.45 &  \emph{25.93} & \bf 24.82 & 21.01 & \bf 18.42  \\
    3.&\ (with finetuning) && 19.47 && \bf 18.68 & &  \\
    4.&\ \emph{(oracle score)} & & & & \emph{24.74} & \\
    \bottomrule
\end{tabularx}
\caption{Perplexity results for ancient history language models (AHLM).}
\label{tab-res}
\end{table*}

\section{Discussion}
The main contributions of ancient history language modeling are twofold.
First, we show that---for at least one popular family of large pretrained LMs---input contexts do not need to be contiguous spans of text. 
These LMs can handle non-contiguous contexts with no additional training. 
This discovery is related to the trend of ``prompt programming'' \citep{shin2020autoprompt,sun2020conditioned,li2021prefixtuning} that has emerged around large, inaccessible models like GPT3 \citep{brown2020language}.
Increasingly, access to such models will be via APIs, with model parameters themselves hidden from client processes.
Prompt programming seeks to discover those inputs which can effectively guide the LM to produce the desired output. 
The selected history format employed in our work can be viewed as a successful input prompt format for document-level language modeling. 

Secondly, we show that changes in probability in a given model can be used as a reward for an auxiliary model to optimize.  
It is not obvious that such perplexity improvement as we observe with our oracle score function would be achievable even within a given textual domain, let alone across domains.
Our work shows that these improvements do generalize. 
Can other tasks benefit from training to optimize this reward?

\section{Related Work}
Prior work has shown the value of extending the context window size for improved language modeling \citep{baevski2019adaptive}.
\citet{sukhbaatar2019adaptive} and \citet{beltagy2020longformer} propose modified attention mechanisms for increasing the effective capacity of the model. 
Alternately, \citet{press2020shortformer} show that staged training and position-infused attention allow language models with shorter context windows to outperform those with longer context windows. 
Cognizant of the expense of training and retraining modern large language models, the current work explores alternative ways of using whatever context window is afforded by a pre-existing model.

Our work is also related to techniques for caching and replaying past experiences. 
\citet{khandelwal2020generalization} retain a datastore of vector representation of training instances. 
At inference, they query the datastore using the current model state to retrieve a collection of next-token experiences. 
They interpolate these experiences with the current model's distribution to improve model perplexity. 
\citet{yogatama2021adaptive} extend this approach with a more sophisticated fusion mechanism. 
Our technique learns a different scoring function from these, one based on the observed performance gain of reusing a past experience. 
Additionally, we bring the various parts of the document history into the pretrained language model's context window, enabling long-distance synthesis of information. 

Similar to our work, \citet{dai-etal-2019-transformer} offer a technique for extending the context window of LMs farther into the document history. 
There, a recurrence mechanism is used to allow information from the previous segment of the document to influence the representation of the next. 
In our method, segments from anywhere in the document history can be brought into the context window. 
Additionally, our technique can piece together disjoint spans from the document history for better coverage. 

Related inference techniques include \citet{krause2018dynamic}, who develop a dynamic evaluation method for adapting model parameters to local context. 
Our method also improves performance at inference, but requires no modification of LM parameters.
As language models grow to hundreds of billions of parameters, we believe our use of training a smaller auxiliary selection function is more sustainable than even shallow retraining. 

Our work fits with recent research into avoiding expensive retraining for large language models. \citet{joshi2020contextualized} inject textual knowledge into a language model's context window to improve question answering. 
Their method uses an entity linker to include additional information about named entities found in the passage. 
Our work learns a selection function from data, and is applicable to passages without named entities.

\citet{shin2020autoprompt} use gradient-guided search to find a collection of keywords to append to an input instance which will improve classification accuracy for that instance. 
Our work deals explicitly with document-level modeling, where context is often too long to apply gradient-based methods. 
\citet{li2021prefixtuning} learn a small space of continuous vectors which are prepended to language model context in a table-to-text task. 
Our technique is complementary to theirs; while learning continuous vectors may also improve our task, the discrete method we propose has the additional advantage of providing interpretable (i.e., textual) explanations for model decisions. 

\section{Conclusion}
We propose a new technique for extending the context available to language models by learning to select useful spans from the out-of-context document history. 
Empirical results across model sizes, training regiments, and datasets demonstrate the wide applicability of the proposed method. 
Importantly, our selection function can improve perplexity with minimal additional training and no modification to the (possibly large) language model parameters themselves.
This function can be trained in one domain and applied in others at inference with good results.
Future work can explore the time and space efficiency of the proposed technique, or explore methods for selecting out of document contexts. 

\bibliographystyle{acl_natbib}
\bibliography{anthology,acl2021}

\begin{thebibliography}{16}
\expandafter\ifx\csname natexlab\endcsname\relax\def\natexlab#1{#1}\fi

\bibitem[{Baevski and Auli(2019)}]{baevski2019adaptive}
Alexei Baevski and Michael Auli. 2019.
\newblock \href {http://arxiv.org/abs/1809.10853} {Adaptive input
  representations for neural language modeling}.

\bibitem[{Beltagy et~al.(2020)Beltagy, Peters, and
  Cohan}]{beltagy2020longformer}
Iz~Beltagy, Matthew~E. Peters, and Arman Cohan. 2020.
\newblock \href {http://arxiv.org/abs/2004.05150} {Longformer: The
  long-document transformer}.

\bibitem[{Brown et~al.(2020)Brown, Mann, Ryder, Subbiah, Kaplan, Dhariwal,
  Neelakantan, Shyam, Sastry, Askell, Agarwal, Herbert-Voss, Krueger, Henighan,
  Child, Ramesh, Ziegler, Wu, Winter, Hesse, Chen, Sigler, Litwin, Gray, Chess,
  Clark, Berner, McCandlish, Radford, Sutskever, and
  Amodei}]{brown2020language}
Tom~B. Brown, Benjamin Mann, Nick Ryder, Melanie Subbiah, Jared Kaplan,
  Prafulla Dhariwal, Arvind Neelakantan, Pranav Shyam, Girish Sastry, Amanda
  Askell, Sandhini Agarwal, Ariel Herbert-Voss, Gretchen Krueger, Tom Henighan,
  Rewon Child, Aditya Ramesh, Daniel~M. Ziegler, Jeffrey Wu, Clemens Winter,
  Christopher Hesse, Mark Chen, Eric Sigler, Mateusz Litwin, Scott Gray,
  Benjamin Chess, Jack Clark, Christopher Berner, Sam McCandlish, Alec Radford,
  Ilya Sutskever, and Dario Amodei. 2020.
\newblock \href {http://arxiv.org/abs/2005.14165} {Language models are few-shot
  learners}.

\bibitem[{Dai et~al.(2019)Dai, Yang, Yang, Carbonell, Le, and
  Salakhutdinov}]{dai-etal-2019-transformer}
Zihang Dai, Zhilin Yang, Yiming Yang, Jaime Carbonell, Quoc Le, and Ruslan
  Salakhutdinov. 2019.
\newblock \href {https://doi.org/10.18653/v1/P19-1285} {Transformer-{XL}:
  Attentive language models beyond a fixed-length context}.
\newblock In \emph{Proceedings of the 57th Annual Meeting of the Association
  for Computational Linguistics}, pages 2978--2988, Florence, Italy.
  Association for Computational Linguistics.

\bibitem[{Feichtenhofer et~al.(2016)Feichtenhofer, Pinz, and
  Zisserman}]{Feichtenhofer2016ConvolutionalTN}
Christoph Feichtenhofer, A.~Pinz, and Andrew Zisserman. 2016.
\newblock Convolutional two-stream network fusion for video action recognition.
\newblock \emph{2016 IEEE Conference on Computer Vision and Pattern Recognition
  (CVPR)}, pages 1933--1941.

\bibitem[{Joshi et~al.(2020)Joshi, Lee, Luan, and
  Toutanova}]{joshi2020contextualized}
Mandar Joshi, Kenton Lee, Yi~Luan, and Kristina Toutanova. 2020.
\newblock \href {http://arxiv.org/abs/2004.12006} {Contextualized
  representations using textual encyclopedic knowledge}.

\bibitem[{Khandelwal et~al.(2020)Khandelwal, Levy, Jurafsky, Zettlemoyer, and
  Lewis}]{khandelwal2020generalization}
Urvashi Khandelwal, Omer Levy, Dan Jurafsky, Luke Zettlemoyer, and Mike Lewis.
  2020.
\newblock \href {http://arxiv.org/abs/1911.00172} {Generalization through
  memorization: Nearest neighbor language models}.

\bibitem[{Krause et~al.(2018)Krause, Kahembwe, Murray, and
  Renals}]{krause2018dynamic}
Ben Krause, Emmanuel Kahembwe, Iain Murray, and Steve Renals. 2018.
\newblock Dynamic evaluation of neural sequence models.
\newblock In \emph{International Conference on Machine Learning}, pages
  2766--2775. PMLR.

\bibitem[{Li and Liang(2021)}]{li2021prefixtuning}
Xiang~Lisa Li and Percy Liang. 2021.
\newblock \href {http://arxiv.org/abs/2101.00190} {Prefix-tuning: Optimizing
  continuous prompts for generation}.

\bibitem[{Lo et~al.(2020)Lo, Wang, Neumann, Kinney, and
  Weld}]{lo-etal-2020-s2orc}
Kyle Lo, Lucy~Lu Wang, Mark Neumann, Rodney Kinney, and Daniel Weld. 2020.
\newblock \href {https://doi.org/10.18653/v1/2020.acl-main.447} {{S}2{ORC}: The
  semantic scholar open research corpus}.
\newblock In \emph{Proceedings of the 58th Annual Meeting of the Association
  for Computational Linguistics}, pages 4969--4983, Online. Association for
  Computational Linguistics.

\bibitem[{Press et~al.(2020)Press, Smith, and Lewis}]{press2020shortformer}
Ofir Press, Noah~A. Smith, and Mike Lewis. 2020.
\newblock \href {http://arxiv.org/abs/2012.15832} {Shortformer: Better language
  modeling using shorter inputs}.

\bibitem[{Radford et~al.(2019)Radford, Wu, Child, Luan, Amodei, and
  Sutskever}]{radford2019language}
Alec Radford, Jeff Wu, Rewon Child, David Luan, Dario Amodei, and Ilya
  Sutskever. 2019.
\newblock Language models are unsupervised multitask learners.

\bibitem[{Shin et~al.(2020)Shin, Razeghi, au2, Wallace, and
  Singh}]{shin2020autoprompt}
Taylor Shin, Yasaman Razeghi, Robert L. Logan~IV au2, Eric Wallace, and Sameer
  Singh. 2020.
\newblock \href {http://arxiv.org/abs/2010.15980} {Autoprompt: Eliciting
  knowledge from language models with automatically generated prompts}.

\bibitem[{Sukhbaatar et~al.(2019)Sukhbaatar, Grave, Bojanowski, and
  Joulin}]{sukhbaatar2019adaptive}
Sainbayar Sukhbaatar, Edouard Grave, Piotr Bojanowski, and Armand Joulin. 2019.
\newblock \href {http://arxiv.org/abs/1905.07799} {Adaptive attention span in
  transformers}.

\bibitem[{Sun and Lai(2020)}]{sun2020conditioned}
Fan-Keng Sun and Cheng-I Lai. 2020.
\newblock \href {http://arxiv.org/abs/2011.07347} {Conditioned natural language
  generation using only unconditioned language model: An exploration}.

\bibitem[{Yogatama et~al.(2021)Yogatama, d'Autume, and
  Kong}]{yogatama2021adaptive}
Dani Yogatama, Cyprien de~Masson d'Autume, and Lingpeng Kong. 2021.
\newblock Adaptive semiparametric language models.
\newblock \emph{arXiv preprint arXiv:2102.02557}.

\end{thebibliography}
\end{document}